\let\accentvec\vec

\documentclass[runningheads]{llncs}

\let\vec\accentvec
\usepackage{amsmath, amssymb} 
\usepackage{booktabs}
\usepackage{bm}
\usepackage{textcomp}
\usepackage{gensymb}
\usepackage{pifont}
\usepackage[T1]{fontenc}
\usepackage{multirow}
\usepackage[rightcaption]{sidecap}
\usepackage{makecell}
\usepackage{xspace}
\usepackage{enumitem}
\usepackage{url}
\usepackage{enumerate}
\usepackage{wrapfig}

\usepackage{graphicx} 
\graphicspath{ {images/} }

\makeatletter
\DeclareRobustCommand\onedot{\futurelet\@let@token\@onedot}
\def\@onedot{\ifx\@let@token.\else.\null\fi\xspace}

\makeatother

\newcommand{\cmark}{\ding{51}}%
\newcommand{\xmark}{\ding{55}}%

\newcommand{\PreserveBackslash}[1]{\let\temp=\\#1\let\\=\temp}
\newcolumntype{C}[1]{>{\PreserveBackslash\centering}p{#1}}
\newcolumntype{R}[1]{>{\PreserveBackslash\raggedleft}p{#1}}
\newcolumntype{L}[1]{>{\PreserveBackslash\raggedright}p{#1}}

\usepackage[bookmarks=false]{hyperref}
\hypersetup{breaklinks=true,colorlinks}

\begin{document}

\pagestyle{headings}
\mainmatter
\title{Simulating Content Consistent Vehicle Datasets with Attribute Descent} 

\authorrunning{Y. Yao et al.}
\titlerunning{Simulating Content Consistent Vehicle Datasets with Attribute Descent}
\author{Yue Yao$^1$ ~ ~ Liang Zheng$^1$~ ~ Xiaodong Yang$^2$~ ~\\ Milind Naphade$^2$~ ~ Tom Gedeon$^1$\\
   {
   }
}
\institute{$^1$Australian National University \\
\email{\{yue.yao, liang.zheng, tom.gedeon\}@anu.edu.au}\\
$^2$NVIDIA \\
\email{yangxd.hust@gmail.com, mnaphade@nvidia.com}
}
\maketitle
\begin{abstract}

This paper uses a graphic engine to simulate a large amount of training data with free annotations. Between synthetic and real data, there is a two-level domain gap, \emph{i.e.,} content level and appearance level. While the latter has been widely studied, we focus on reducing the content gap in attributes like illumination and viewpoint. To reduce the problem complexity, we choose a smaller and more controllable application, vehicle re-identification (re-ID). We introduce a large-scale synthetic dataset VehicleX. Created in Unity, it contains 1,362 vehicles of various 3D models with fully editable attributes. We propose an attribute descent approach to let VehicleX approximate the attributes in real-world datasets. Specifically, we manipulate each attribute in VehicleX, aiming to minimize the discrepancy between VehicleX and real data in terms of the Fr\'{e}chet Inception Distance (FID). This attribute descent algorithm allows content domain adaptation (DA) orthogonal to existing appearance DA methods. We mix the optimized VehicleX data with real-world vehicle re-ID datasets, and observe consistent improvement. With the augmented datasets, we report competitive accuracy. We make the dataset, engine and our codes available at \url{https://github.com/yorkeyao/VehicleX}.

\keywords{vehicle retrieval, domain adaptation, synthetic data}

\end{abstract}

\section{Introduction}

Data synthesis, as can be conveniently performed in graphic engines, provides valuable convenience and flexibility for the computer vision area~\cite{richter2016playing,sakaridis2018semantic,ruiz2019learning,tremblay2018training,sun2019dissecting}. One can synthesize a large amount of training data under various combinations of environmental factors even from a small number of 3D object/scene models. However, there exists a huge domain gap between synthetic data and real-world data~\cite{kar2019meta,ruiz2019learning}. In order to effectively alleviate such a domain gap, it should be addressed from two levels: \textbf{content level} and \textbf{appearance level}~\cite{kar2019meta}. While much existing work focuses on appearance level domain adaptation~\cite{deng2018image,hoffman2018cycada,zhong2018camera}, we focus on the content level, \emph{i.e.,} learning to synthesise data with similar content to the real data, as different computer vision tasks require different image contents.

Our system is designed based on the following considerations. It is expensive to collect large-scale real-world datasets for muti-camera system like re-ID. During annotation, one needs to associate an object across different cameras, which is a difficult and laborious process as objects might exhibit very different appearances in different cameras. In addition, there also has been an increasing concern over privacy and data security, which makes collection of large real datasets difficult~\cite{ristani2016MTMC,yao2020information}. On the other hand, we can see that datasets can be very different in their content. Here content means the object layout, illumination, and background in the image. For example, the VehicleID dataset~\cite{liu2016deep} consists mostly of car rears and car fronts, while vehicle viewpoints in the VeRi-776 dataset~\cite{liu2016large} cover a very diverse range. Though the VehicleID dataset has a large number of identities which is useful for model training, this content-level domain gap might cause a model trained on VehicleID to have poor performance on VeRi. Most existing domain adaptation methods work on the pixel level or the feature level so as to allow the source and target domains to have similar appearance or feature distributions. However, these approaches are not capable of handling content differences, as can often be encountered when training on synthetic data and testing on real data. 

\begin{figure}[t]
\begin{center}
	\includegraphics[width=1\linewidth]{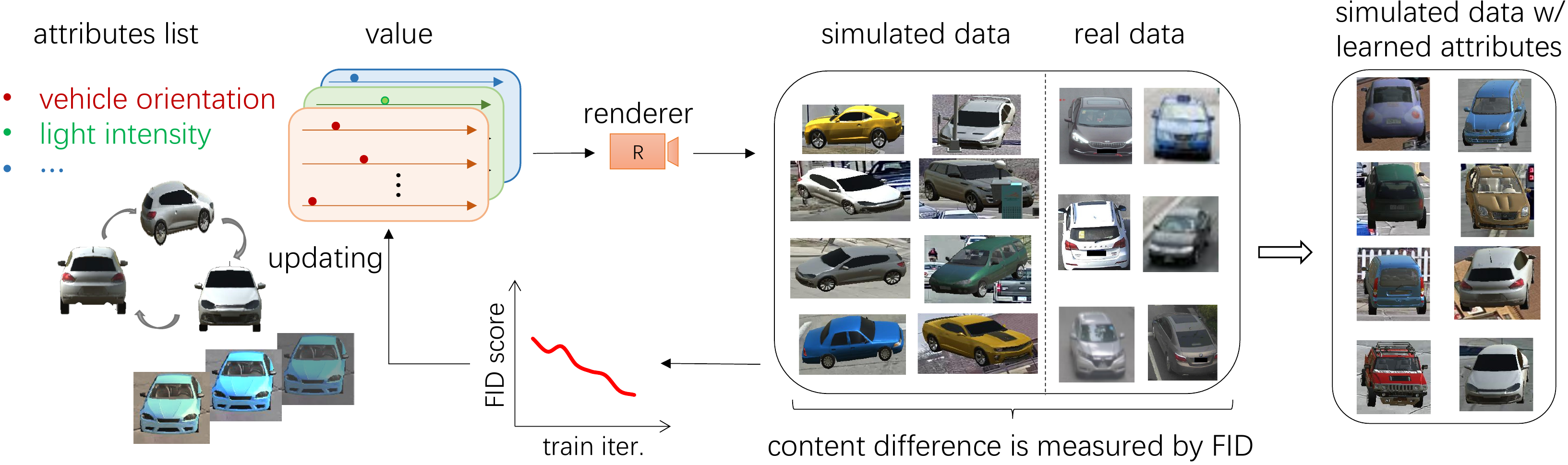}
\end{center}
\caption{
	System workflow. (\textbf{Left:}) given a list of attributes and their values, we use a renderer (\emph{i.e.,} Unity) for vehicle simulation. We compute the Fr\'{e}chet Inception Distance (FID) between the synthetic and real vehicles to indicate their distribution difference. By updating the values of attributes using the proposed attribute descent algorithm, we can minimize FID along the training iterations. (\textbf{Right:}) we use the learned attributes values that minimize FID to generate synthetic data to be used for re-ID model training.
}
\label{fig:system_flow}
\end{figure}

Based on above considerations, we aim to utilize flexible 3D graphic engine to
1) scale up the real-world training data without labeling and privacy concerns, and 2) build synthetic data with \emph{less content domain gap} to real-world data. To this end, we make contributions from two aspects. 
First, we introduce a large-scale synthetic dataset named VehicleX, which lays the foundation of our work. It contains 272 backbone models, with different colored textures, and creates 1,362 different vehicles. Similar to many existing 3D synthetic datasets such as PersonX~\cite{sun2019dissecting} and ShapeNet~\cite{chang2015shapenet}, VehicleX has editable attributes and is able to generate a large training set by varying object and environment attributes. 
Second, based on the VehicleX, we propose an attribute descent method which automatically configures the platform attributes, such that the synthetic data shares similar content distributions with the real data of interest. As shown in Figure~\ref{fig:system_flow}, specifically, we manipulate the range of five key attributes closely related to the real dataset content. To measure the distribution discrepancy between the synthetic and real data, we use the FID score and aim to minimize it. In each epoch, we optimize the values of attributes in a specific sequence.

We show the effectiveness of attribute descent by training with VehicleX only and joint training with real-world datasets. The synthetic training data with optimized attributes can improve re-ID accuracy under both settings. 
Furthermore, under our joint training scheme, with VehicleX data, we achieve competitive re-ID accuracy with the state-of-the-art approaches, validating the effectiveness of learning from synthetic data. A subset of VehicleX has been used in the 4th AICITY challenge~\cite{naphade20204th}.\footnote{\url{https://www.aicitychallenge.org/}}

\section{Related Work}

\textbf{Vehicle re-identification} has received increasing attention in the past few years, and many effective systems have been proposed~\cite{khorramshahi2019dual,wang2017orientation,tang2019pamtri,zhou2018aware}, generally with specially designed or fine-tuned architectures. 
In this paper, our baseline system is built with commonly used loss functions~\cite{zheng2016mars,hermans2017defense,szegedy2016rethinking} with no bells and whistles. Depending on the camera conditions, location and environment, existing vehicle re-ID datasets usually have their own distinct characteristics. For example, images in the VehicleID~\cite{liu2016deep} are either captured from the car front or the back. In comparison, the VeRi-776~\cite{liu2016large} includes a wider range of viewpoints. 
The recently introduced CityFlow~\cite{tang2019cityflow} has distinct camera heights and backgrounds. Apart from dataset differences, there also exists huge differences between cameras in a single dataset~\cite{zhong2018camstyle}. For example, a camera filming a crossroad naturally has more vehicles orientation than a camera on a straight road.
Because of these characteristics, within a specific dataset, we learn attributes for each camera and simulate that filming environment in a 3D engine. As a result, our proposed data simulation approach will make synthetic data more similar to the real-world in key attributes, and thus can effectively augment re-ID datasets due to its strong ability in content adaptation.



\textbf{Appearance(style)-level domain adaptation.} Domain adaptation is often used to reduce the domain gaps between the distributions of two datasets. Till now, the majority of work in this field focuses on discrepancies in image style, such as real vs. synthetic~\cite{bak2018domain} and real vs. sketch~\cite{peng2019moment}. For example, some use the cycle generative adversarial network (CycleGAN) to reduce the style gap between two domains~\cite{hoffman2018cycada,shrivastava2017learning,deng2018image}, as well as various constraints being exerted on the generative model such that useful properties are preserved. 
While these works have been shown to be effective in reducing the style domain gap, a fundamental problem remains to be solved, \emph{i.e.}, the content difference. 


\begin{figure}[t] 
    \centering
    \begin{center}
        \includegraphics[width=1\linewidth]{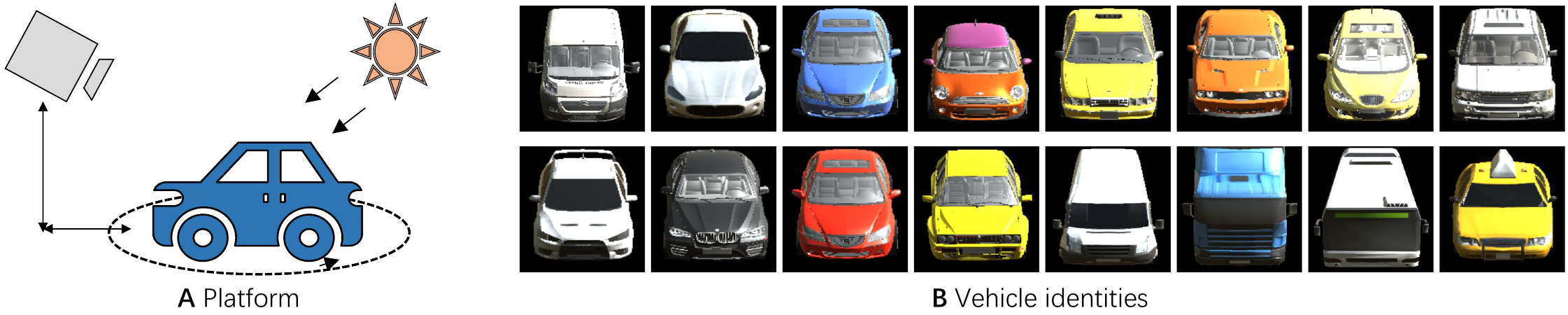}
        \caption{The VehicleX engine. (A) An illustration of the rendering platform. We adjust the vehicle orientation, light direction and intensity, camera height, and the distance between the camera and the vehicle. (B) 16 different vehicle identities are shown. 
        }
        \label{fig: Platform}
    \end{center}
\end{figure}

\textbf{Content-level domain adaptation}, to our knowledge, has been discussed by only a few existing works~\cite{kar2019meta,ruiz2019learning}. For~\cite{ruiz2019learning}, their main contribution is clever usage of the task loss to guide the domain adaptation procedure. But for the re-ID task, we will search attributes for each camera but get task loss across camera systems. That is, task loss can only be gotten when all camera attributes are set. As a result, it is hard to optimise attributes for a single camera using loss from a cross-camera system. For~\cite{kar2019meta}, they use Graph Convolution Neural Network (GCN) to optimise the probability grammar for scene generation (\emph{e.g.,} detection task). Their target is to solve the relationship between multiple objects. But in re-ID settings, we only have one object (car) to optimize. As their method cannot be directly used for the re-ID task, we adopt their advantages and make new contributions. On the one hand, we adopt the idea of Ruiz \emph{et al.}~\cite{ruiz2019learning} that represents attributes using predefined distributions. We are also motivated by Kar \emph{et al.}~\cite{kar2019meta}, who suggest that some GAN evaluation metrics (\emph{e.g.,} KID~\cite{binkowski2018demystifying}) are potentially useful to measure content differences. In practice, we propose attribute descent, which does not involve random variables and has easy-to-configure step sizes. 






\textbf{Learning from 3D simulation.} Due to low data acquisition costs, learning from 3D world is an attractive way to increase training set scale. But unlike other synthetic data (\emph{e.g.,} images generated by GAN~\cite{zheng2017unlabeled}), 3D simulation provides more accurate data labeling, flexibility in content generation and scalability in resolution, as GAN generated image may suffers from these problems. In the 3D simulation area, many applications exist in areas such as semantic segmentation~\cite{hoffman2018cycada,gaidon2016virtual,xue2020learning}, navigation~\cite{kolve2017ai2}, detection~\cite{kar2019meta,hou2020multiview}, object re-identification~\cite{sun2019dissecting,tang2019pamtri}, \emph{etc}. Usually, prior knowledge is utilized during data synthesis since we will inevitably need to determine the distribution of attributes in our defined environment. Tremblay \emph{et al.} suggest that attribute randomness in a reasonable range is beneficial~\cite{tremblay2018training}. Even if it is random, we need to specify the range of random variables in advance. Our work investigates and learns these attribute distributions for vehicle re-ID.



\section{VehicleX Engine}
We introduce a large-scale synthetic dataset generator named VehicleX that includes three components: (1) vehicles rendered using the graphics engine Unity, (2) a Python API that interacts with the Unity 3D engine, and (3) detailed labels including car type and color. 

VehicleX has \textbf{a diverse range of realistic backbone models and textures}, allowing it to be able to adapt to the variance of real-world datasets. It has 
272 backbones that are hand-crafted by professional 3D modelers. The backbones include ten mainstream vehicle types including sedan, SUV, van, hatchback, MPV, pickup, bus, truck, estate, sportscar and RV. Each backbone represents a real-world model. From these backbones, we obtain 1,362 variances (\emph{i.e.,} identities) by adding various colored textures or accessories. A comparison of VehicleX with some existing vehicle re-ID datasets is presented in Table~\ref{table:Datasets}. VehicleX is three times larger than the synthetic PAMTRI dataset~\cite{tang2019pamtri} in identities, and can potentially render an unlimited number of images from various attributes.
In experiments, we will show that our VehicleX benefits real-world testing either when used alone or in conjunction with a real-world training set. 

In this work, VehicleX can be set to training mode and testing mode. In training mode, VehicleX will render images with black background and these images will be used for attribute descent (see Section~\ref{sec:attribute_descent}); in comparison, the testing mode uses random images (\emph{e.g.,} from CityFlow~\cite{tang2019cityflow}) as backgrounds, and generates attribute-adjusted images. In addition, to increase randomness and diversity, the testing mode contains random street objects such as lamp posts, billboards and trash cans. Figure~\ref{fig: Platform} shows the simulation platform, and some sample vehicle identities. 

\begin{table}[t]\footnotesize
\caption{Comparison of some real-world and synthetic vehicle re-ID datasets. "Attr" denotes whether the dataset has attribute labels (\emph{e.g.,} orientation). Our identities are different 3D models, thus can potentially render an unlimited number of images under different environment and camera settings. VehicleX is released open source and can be used to generate (possess) an unlimited number of images (cameras). 
		\label{table:Datasets}}
\begin{center}
	\setlength{\tabcolsep}{3.1mm}{
		\begin{tabular}{c|l|c|c|c|c} 
			\Xhline{1.2pt}
			\multicolumn{2}{c|}{Datasets}	&  \#IDs & \multicolumn{1}{c|}{\#Images} & \#Cameras  & \# Attr \footnotesize \\
			\hline 
			\multirow{4}{*}{real}	& VehicleID~\cite{liu2016deep} &26,328  &222,629 &2 & \xmark \\
			&CompCar~\cite{yang2015large} &4,701 &136,726 & - & \xmark \\
			&VeRi-776~\cite{liu2016large} & 776 & 49,357  & 20 &\cmark  \\
			&CityFlow~\cite{tang2019cityflow} & 666 &  56,277 & 40 & \xmark \\
			\hline
			\multirow{2}{*}{{synthetic}}	&PAMTRI~\cite{tang2019pamtri}  & 402 & 41,000 & - & \cmark \\
			& VehicleX  & 1,362 & $\infty$  & $\infty$ & \cmark \\ 
	
			\Xhline{1.2pt} 			
	\end{tabular}}
\end{center}
\end{table}

We build the \textbf{Unity-Python interface} using the Unity ML-Agents toolkit~\cite{juliani2018unity}. It allows Python to modify the attributes of the environment and vehicles, and obtain the rendered images. 
With this API, given the attributes needed, users can easily obtain rendered images without expert knowledge about Unity. The code of this API is released together with VehicleX. 

VehicleX is a large scale public 3D vehicle dataset, with real-world vehicle types. We focus on vehicle re-ID task in this paper but our proposed 3D vehicle models also has potential benefits for many other tasks, such as semantic segmentation, object detection, fine-grained classification, 3D generation or reconstruction. It gives flexibility to computer vision systems to freely edit the content of the object, thus enabling new research in content-level image analysis. 

\begin{figure}[t] 
    \centering
    \begin{center}
        \includegraphics[width=1\linewidth]{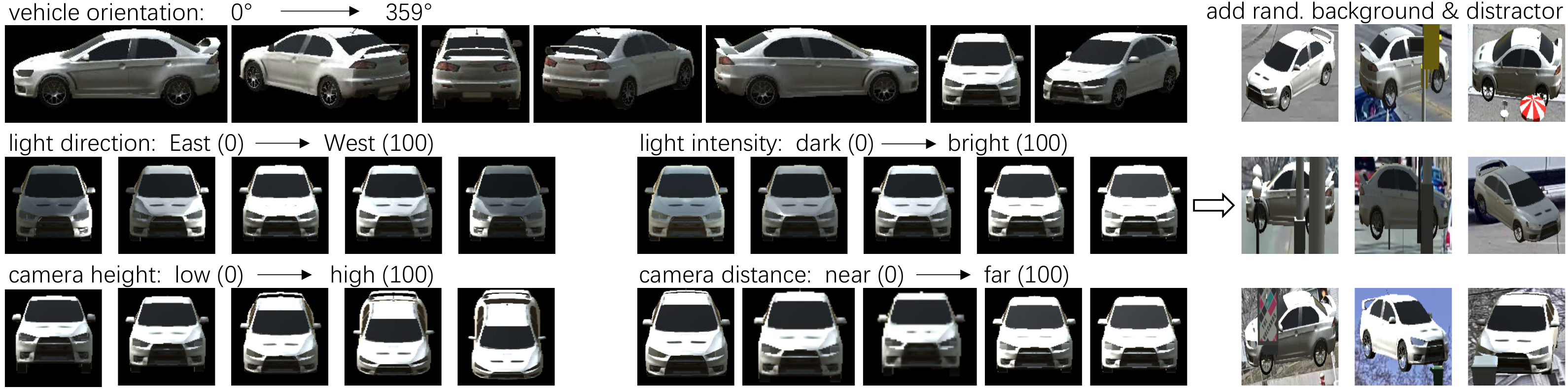}
        \caption{(\textbf{Left:}) Attribute editing. We rotate the vehicle, edit light direction and intensity, or change the camera height and distance. 
Numbers in the bracket correspond to the attribute values in Unity. (\textbf{Right:}) We further add random backgrounds and distractors to the attribute-adjusted vehicles when they are used in the re-ID model. 
        }
        \label{fig: editing}
    \end{center}
\end{figure}

\section{Proposed Method}\label{sec:attribute_descent}
\subsection{Attribute Distribution Modeling}\label{sec:att_model}
\textbf{Important attributes.}\label{sec:attributes} 
For vehicle re-ID, we consider the following attributes to be potentially influential on the training set simulation and testing accuracy. Figure~\ref{fig: editing} shows examples of the attribute editing process. 

\begin{itemize}[noitemsep,topsep=0pt]
	\item \textbf{Vehicle orientation} is the horizontal viewpoint of a vehicle and takes a value between 0\degree and 359\degree. In the real world, this attribute is important because the camera position is usually fixed and vehicles usually move along predefined trajectories. Therefore, the distribute of vehicle orientation of real world dataset is usually multimodal and tend to exhibit certain patterns under a certain camera view. 
	\item \textbf{Light direction} simulates daylight as cars are generally presented in outdoor scenes.
	Here, we assume directional parallel light, and the light direction is modeled from east to west, which is the movement trajectory of the sun. 
	\item \textbf{Light intensity} is usually considered a critical factor for re-ID tasks. Factors include glass refection and shadows will seriously influence the results. We manually defined a reasonable range for intensity from dark to light. 
	\item \textbf{Camera height} describes the vertical distance from the ground, and significantly influences viewpoints. 
	\item \textbf{Camera distance} determines the horizontal distance from vehicles. This factor has a strong effect on the vehicle resolution since the resolution of the entire image is predefined as 1920$\times$1080. Additionally, the distance has slight impacts on viewpoints.
\end{itemize}

\begin{figure}[t] 
    \centering
    \begin{center}
        \includegraphics[width=1\linewidth]{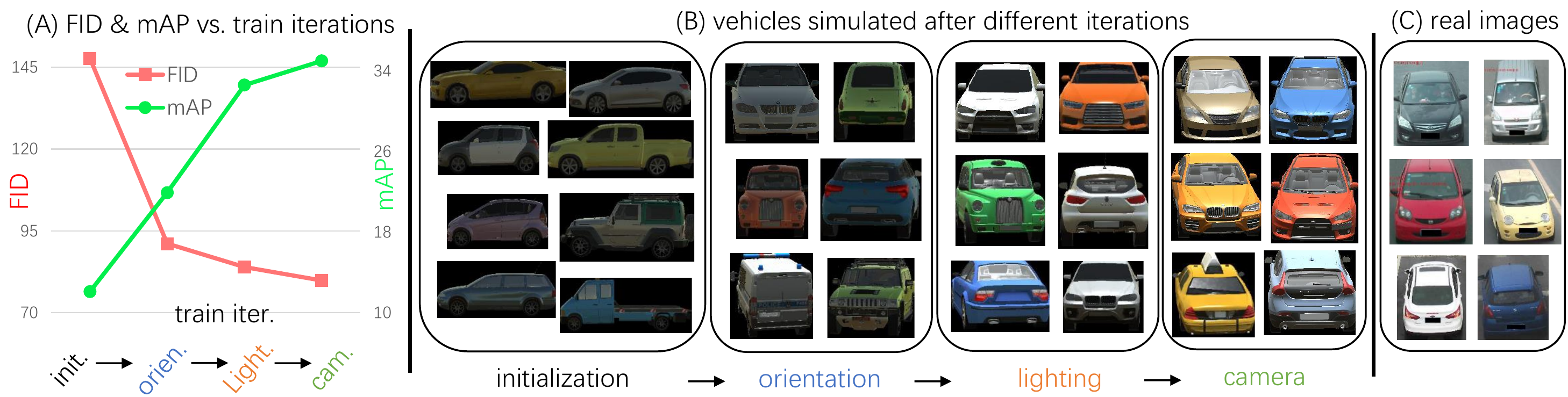}
        \caption{Attribute descent visualization on the VehicleID~\cite{liu2016deep}. (A) The FID-mAP curve through training iterations. The FID successively drops (lower is better) and domain adaptation mAP successively increases (higher is better) during attribute descent. For illustration simplicity, we use ``light'' to denote light direction and intensity, and use ``cam.'' to denote camera height and distance. (B) We show the synthetic vehicles in each iteration. We initialize the attributes by setting orientation to right, the light intensity to dark, light direction to west, camera height to being equal to the vehicle, and camera distance to medium. The content of those images become more and more similar to (C) the target real images through the optimization procedure.
        }
        \label{fig:training}
    \end{center}
\end{figure}

\textbf{Distribution modeling.} 
We model the aforementioned attributes with single Gaussian distributions or Gaussian Mixture Model (GMM). This modeling strategy is also used in Ruiz \emph{et al.}'s work~\cite{ruiz2019learning}.
We denote the attribute list as: $ \mathcal{A} = (a_1, a_2,...,a_N)$, where $N$ is the number of attributes considered in the system, and $a_i, i = 1,...,N$ is the random variable representing the $i$th attribute. 

For the vehicle orientation, we use a GMM to capture its distribution. This is based on our prior knowledge that the field-of-view of a camera covers either a road or an intersection. If we do not consider vehicle turning, there are rarely more than four major directions at a crossroad. In this work, we set a GMM with 6 components.
For lighting conditions and camera attributes, we use four independent Gaussian distributions. 
Therefore, given $N$ attributes, we optimize $M$ mean values of the Gaussians, where $M\geq N$.

We speculate that the means of the Gaussian distributions or components are more important than the standard deviations because means reflect how the majority of the vehicles look. Although our method has the ability to handle variances, this would significantly increase the search space. As such, we predefine the values of standard deviations
and only optimize the means of all the Gaussians $\bm{\mu}= (\mu_1, \mu_2, ..., \mu_M)$, where $\mu_i\in \mathbb{R}, i=1,...,M$ is the mean of the $m$th Gaussian. As a result, given the means $\bm{\mu}$ of the Gaussians, we can sample an attribute list as $\mathcal{A} \sim G(\bm{\mu})$, where G is a function that generates a set of attributes given means of Gaussian. 

\subsection{Optimization}
The objective of our optimization is to train a model to generate a dataset that has a similar content distribution with respect to a target real dataset. 

\textbf{Measuring distribution difference.} 
We need to precisely define the distribution difference before we apply any optimization algorithm. There potentially exists two directions: using the appearance difference, and the task loss on the validation set. But as re-ID is a cross-camera task, it is indirect and difficult for us to optimise attributes for a single camera using loss from a cross-camera system. So we focus on the appearance difference. For the appearance difference,
we use the Fr\'{e}chet Inception Distance (FID)~\cite{heusel2017gans} to quantitatively measure the distribution difference between two datasets.
Adversarial loss is not used as the measurement directly since there exists a huge appearance difference between synthetic and real data, and the discriminator would easily detect the specific detailed differences between real and generated, and yet not be useful. 

Formally, we denote the sets of synthetic data and real data as $X_s$ and $X_r$ respectively, 
where $X_s = \{ \mathcal{R}(\mathcal{A}_1), \cdots, \mathcal{R}(\mathcal{A}_K) | \mathcal{A}_k \sim G(\bm{\mu}) \}$, and $\mathcal{R}$ 
is our rendering function through the 3D graphics engine working on a given attribute list $\mathcal{A}$ that controls the environment. $K$ is the number of images in the synthetic dataset. For the FID calculation, we employ the Inception-V3 network~\cite{szegedy2016rethinking} to map an image into its feature space. We view the feature as a multivariate real-valued random variable and assume that it follows a Gaussian distribution. 
To measure the distribution difference between two Gaussians, we resort to their means and covariance matrices. Under FID, the distribution difference between synthetic data and real data is written as, 
\begin{equation}
\begin{split}
\mbox{FID}(X_s, X_r) = \left \| \bm{\mu}_s - \bm{\mu}_r  \right \|^{2}_{2} + 
         Tr(\bm{\Sigma}_s + \bm{\Sigma}_r -2 (\bm{\Sigma}_s \bm{\Sigma}_r)^{\frac{1}{2}}),
\end{split}
\label{eq:fid}
\end{equation}
where $\bm{\mu}_s$ and $\bm{\Sigma}_s$ denote the mean and covariance matrix of the feature distribution of the synthetic data, and $\bm{\mu}_r$ and $\bm{\Sigma}_r$ are from the real data. 


\textbf{Attribute descent.}
An important difficulty for attribute optimization is that the rendering function (through the 3D engine Unity) is not differentiable, so the widely used gradient-descent based methods cannot be readily used. 
Under this situation, there exist several methods for gradient estimation, such as finite-difference~\cite{kar2019meta} and reinforcement learning~\cite{ruiz2019learning}. 
However, these methods are developed in scenarios where there are many parameters to optimize. In comparison, our system only contains a few parameters, allowing us to design a more stable and efficient approach that is sufficiently effective in finding a close to global minimum. 

We are motivated by coordinate descent, an optimization algorithm that can work in derivative-free contexts~\cite{wright2015coordinate}. The most commonly known algorithm that uses coordinate descent is $k$-means~\cite{lloyd1982least}. 
Coordinate descent successively minimizes along coordinate directions to find a minimum of a function. The algorithm selects a coordinate to perform the search at each iteration. Compared with grid search, coordinate descent significantly reduces the search time, based on the hypothesis that each parameter is relatively independent. For our designed attributes, we study their independence in subsection~\ref{sec:single_cam}. 

Using Eq.~\ref{eq:fid} as the objective function, we propose attribute descent to optimize each single attribute in the attribute list. Specifically, we view each attribute as a coordinate in the coordinate descent algorithm. In each iteration, we successively change the value of an attribute to search for the minimum value of the objective function. Formally, for our defined parameters $\bm{\mu}$ for attributes list $\mathcal{A}$, the objective is to find 
\begin{equation}
\begin{split}
\bm{\mu} = \mathop{\arg\min}_{\bm{\mu}} \mbox{FID}(X_s, X_r),\qquad\\
X_s = \{ \mathcal{R}(\mathcal{A}_1), \cdots, \mathcal{R}(\mathcal{A}_K) | \mathcal{A}_k \sim G(\bm{\mu}) \}.
\end{split}
\end{equation}
We achieve this objective iteratively. Initially, we have
\begin{equation}
\bm{\mu}^{0} = (\mu_{1}^{0}, \cdots, \mu_{M}^{0}),
\end{equation}
At epoch $j$, we optimize a single variable $\mu_{i}^{j}$ in $\bm{\mu}$,
\begin{equation}
\begin{split}
\mu_{i}^{j} = \mathop{\arg\min}_{z \in S_{i}} \mbox{FID}(X_s, X_r),\qquad\\
X_s = \{ \mathcal{R}(\mathcal{A}_1), \cdots, \mathcal{R}(\mathcal{A}_K) |  \mathcal{A}_k \sim G(\mu_1^{j}, \\ 
\cdots, \mu_{i-1}^{j}, z, \mu_{i+1}^{j-1}, \cdots, \mu_M^{j-1}) \},
\end{split}
\end{equation}
where 
the $S_i, i=1,...,M$ define a specific search space for mean variable $\mu_i$. For example, the search space for vehicle orientation is from $0^{\circ}$ to $330^{\circ}$ by $30^{\circ}$ degree increments; the search space for camera height is the equally divided editable range with 9 segments. $j = 1,\cdots, J$ are the training epochs. One epoch is defined as all attributes being updated once. 
In this algorithm, we perform greedy search for the optimized value of an attribute in each iteration, and achieve a local minimum for each attribute when fixing the rest. 

\textbf{Discussion.} 
In Section~\ref{sec:single_cam} we show that attribute descent (non-gradient solution) is superior to our implementation of reinforcement learning (gradient-based solution). Attribute descent, inherited from the coordinate descent algorithm, is simple to implement and steadily leads to convergence. It is a new optimization tool in the learning to synthesize literature and avoids drawbacks such as difficulty in optimization and sensitivity to hyper-parameters. 
That being said, we note that our method is effective in small-scale environments like vehicle bounding boxes where only a small number of attributes need to be optimized. In more complex environments, we suspect that reinforcement learning algorithms should also be effective. 



\begin{figure}[t]  
    \begin{center}
        \includegraphics[width=4.0in]{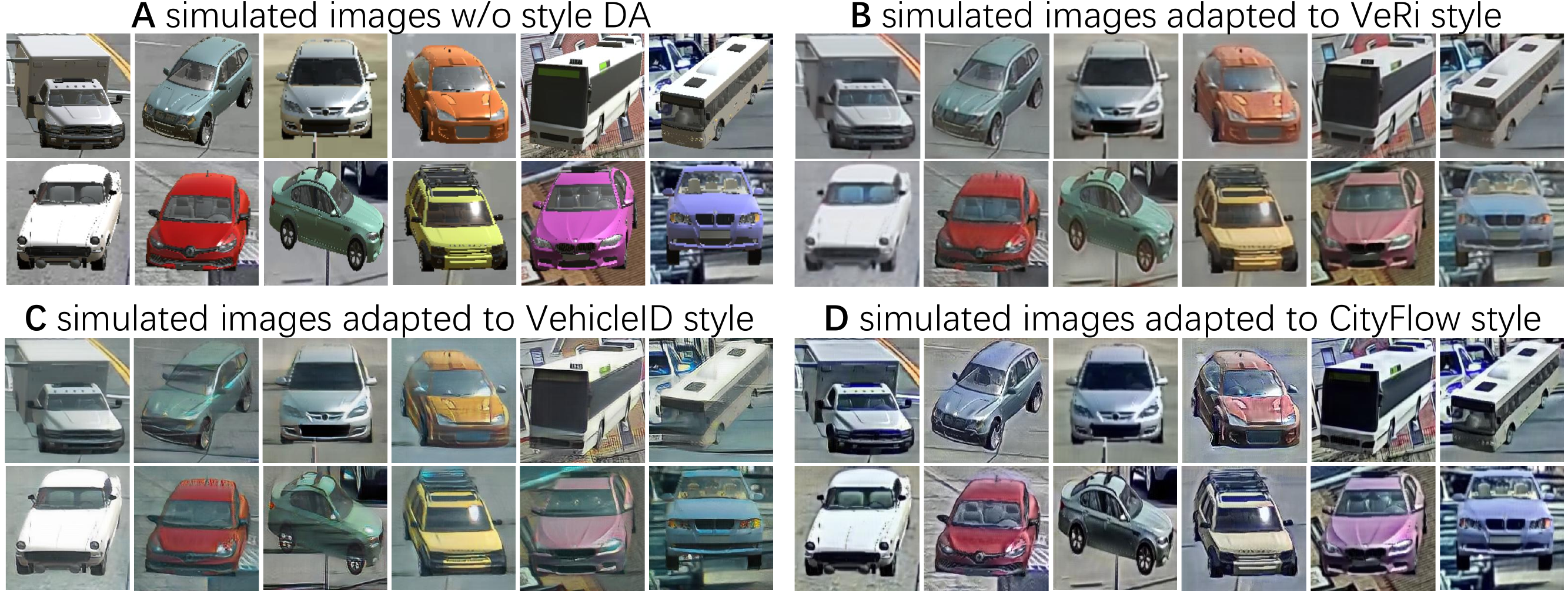}
        \caption{Images w/ and w/o style domain adaptation. (A) Synthetic images without style domain adaptation. (B)(C)(D) We translate images in (A) to the style of VeRi, VehicleID and CityFlow, respectively, using SPGAN~\cite{deng2018image}.}
        \label{figure:SPGAN}
    \end{center}
\end{figure}


\section{Experiment}

\subsection{Datasets and Evaluation Protocol}
We use three real-world datasets for evaluation. \textbf{VehicleID}~\cite{liu2016deep} is at a large scale, containing 222,629 images of 26,328 identities. Half of the identities are used for training, and the other half for testing. Officially there are 3 test splits.
The \textbf{VeRi-776} dataset~\cite{liu2016large} contains 49,357 images of 776 vehicles captured by 20 cameras. The vehicle viewpoints and illumination cover a diverse range. The training set has 37,778 images, corresponding to 576 identities; the test set has 11,579 images of 200 identities. There are 1,678 query images. The train / test sets share the same 20 cameras. \textbf{CityFlow}~\cite{tang2019cityflow} has more complex environments, and it has 40 cameras in a diverse environment where 34 are used in the training set. The dataset has in total 666 IDs where half are used for training and the rest for testing.
We use mean average precision (mAP) and Rank-1 accuracy to measure the re-ID performance. 



\subsection{Implementation Details}
\textbf{Data generation.} For the VehicleID dataset,
we only optimize a single attribute list targeting the VehicleID training set. 
But most re-ID datasets like VeRi-776 and CityFlow are naturally divided according to multiple camera views. Since a specific camera view usually has stable attribute features (\emph{e.g.,} viewpoint), we perform the proposed attribute descent algorithm on each individual camera, so as to simulate images with similar content to images from each camera. For example, we optimize 20 attribute lists using the VeRi-776 training set, which has 20 cameras. Attribute descent is performed for two epochs.
\begin{wraptable}{r}{4cm}
\caption{Re-ID accuracy (mAP) w/ and w/o style DA when training with synthetic data only. We clearly observe style DA brings significant improvement and thus is necessary.}
\centering
\resizebox{3.4cm}{0.6cm}{
\begin{tabular}{c|cc} 
\Xhline{1.2pt}        StyleDA         & VehicleID & VeRi    \\ 
\hline
\xmark & 24.36     & 12.35      \\
\cmark & \textbf{35.33}     & \textbf{21.29}    \\
\Xhline{1.2pt}
\end{tabular}}
\label{table:styleDA}
\end{wraptable}
One epoch is defined as all attributes in the list being updated once.


\textbf{Image style transformation.} We apply SPGAN~\cite{deng2018image} for image style transformation, which is a state-of-the-art algorithm in style level domain adaptive re-ID. Sample results are shown in Figure~\ref{figure:SPGAN} and influence is shown in Table~\ref{table:styleDA}.
Image translation models are trained using 112,042 images with random attributes as source domain and the training set in three vehicle datasets as target domain separately. When performing SPGAN for learned attributes data, we directly inference the learned attributes images, based on the fact that our learned attributes are a subset of the random range.

\textbf{Baseline configuration.} 
For VeRi and VehicleID, we use the ID-discriminative embedding (IDE)~\cite{zheng2016mars}. We adopt the strategy from~\cite{luo2019bag} which adds batch normalization and removes ReLU after the final feature layer. 
We also use the part-based convolution baseline (PCB)~\cite{sun2018beyond} on VeRi for improved accuracy. In PCB, we horizontally divide the picture into six equal parts and perform classification on each part. For CityFlow training, we use the setting from~\cite{luo2019bag} using a combination of the cross-entropy loss and the triplet loss.

\setlength{\tabcolsep}{3.3mm}
\begin{table*}[t]
\small
\caption{Method comparison on VehicleID in data augmentation. Our method is built on IDE~\cite{zheng2016mars} with the cross-entropy (CE) loss. Attribute descent consistently improves over both the baseline and random attributes, and is competitive compared with the state-of-the-art. ``R'' means training use real data only. ``R+S'' denotes that both synthetic data and real data are used in training. ``Small'', ``Medium'' and ``Large'' refers to the number of vehicles on the VehicleID test set~\cite{liu2016deep}.}
\resizebox{1\textwidth}{!}{%
\begin{tabular}{p{2cm}|p{0.6cm}<{\centering}|ccc|ccc|ccc}
\Xhline{1.2pt}
\multirow{2}{*}{Method} &   \multirow{2}{*}{Data}     &   &   Small  &        &  &  Medium   &        &  &  Large &    \\ \cline{3-11} 
                        & &  Rank-1 & Rank-5 & mAP & Rank-1 & Rank-5 & mAP & Rank-1 & Rank-5 & mAP \\ \hline
   RAM~\cite{liu2018ram}    & R & 75.2  &  91.5      &  -   &   72.3     & 87.0       &  -  &    67.7   &   84.5    &  -   \\
AAVER~\cite{khorramshahi2019dual}  &  R    &   74.69     &  93.82   &  -    &    68.62    &   89.95 &  -   &  63.54  & 85.64   &  -   \\
 GSTE~\cite{bai2018group}    & R &  75.9  &  84.2  &  75.4   &   74.8    &    83.6    &  74.3    &    74.0    &   82.7    &  72.4   \\
 \hline
IDE (CE loss)       &   R  &  77.35     &   90.28    &   83.10  &     75.24   &    87.45    &  80.73   &    72.78    &  85.56     &  78.51   \\
Ran. Attr.    & R+S   &   80.2   &   93.98    & 85.95    &   76.94     &   90.84     &   82.67  &   73.45     & 88.66  &  80.55   \\
Attr. Desc.    & R+S  &     \textbf{81.50}    &    \textbf{94.85}     &   \textbf{87.33}    &    \textbf{77.62}  &     \textbf{92.20}   &   \textbf{83.88}  &    \textbf{74.87}    &    \textbf{89.90}   &  \textbf{81.35}   
\\\Xhline{1.2pt}
\end{tabular}}
\label{table:comparison-VID}
\end{table*}

\begin{table}[t]
\centering
\caption{FID values between the generated data and VehicleID after Epoch I and II (attribute descent is performed for two epochs). Different orders of attributes are tested. 
`C', `O' and `L' refer to camera, orientation and lighting, respectively. After Epoch II, the FID values are generally similar, suggesting that the correlation among attributes is weak, and so they are mostly independent.} 
\resizebox{1\textwidth}{!}{
\begin{tabular}{l|cccccc} 
\Xhline{1.2pt}
          & C $\rightarrow$ O $\rightarrow$ L  & C $\rightarrow$ L $\rightarrow$ O  & O $\rightarrow$ L $\rightarrow$ C  & O $\rightarrow$ C $\rightarrow$ L  & L $\rightarrow$ C $\rightarrow$ O  & L $\rightarrow$ O $\rightarrow$ C   \\ 
\hline
FID (Epoch I)  &          98.38                          &        99.57                            &      78.67                              &        80.94                            &              104.84                      &           81.20                          \\
FID (Epoch II) &             78.42                       &        77.18                            &                77.96                    &             79.54                       &                   78.48                 &             77.06                        \\ \hline
\Xhline{1.2pt}
\end{tabular}}
\label{table:attr_dep}
\end{table}

\textbf{Experiment protocol.} We evaluate our method on both vehicleX training and joint training settings. Under vehicleX training, we train our model on VehicleX and test on real world data. Under joint training, we combine the VehicleX data and real world data and perform two-stage  training; testing is
 \begin{wraptable}{r}{5.2cm}
\caption{Comparison of the Re-ID accuracy (mAP) of two stage training when performing joint training. We can see a significant performance boost from Stage \uppercase\expandafter{\romannumeral1} to Stage \uppercase\expandafter{\romannumeral2}. }
\centering
\resizebox{5cm}{0.6cm}{
\begin{tabular}{p{1.2cm}|p{1cm}<{\centering}p{0.7cm}<{\centering}p{1.2cm}<{\centering}} 
\Xhline{1.2pt}                  & VehicleID & VeRi  & CityFlow  \\ 
\hline
Stage \uppercase\expandafter{\romannumeral1} & 77.54     & 69.39 & 33.54     \\
Stage \uppercase\expandafter{\romannumeral2} & \textbf{81.35}     & \textbf{70.62} & \textbf{37.16}     \\
\Xhline{1.2pt}
\end{tabular}}
\label{table:two-stage}
\end{wraptable}
 on the same real-world data. 

\textbf{Two-stage training} is conducted in joint training with three real-world datasets~\cite{zheng2019vehiclenet}. We mix synthetic dataset and a real-world dataset in the first stage and finetune on the real-world dataset only in the second stage. Taking CityFlow for example, in the first stage, we train on both real and synthetic data. We classify vehicle images into one of the 1,695 (333 from real + 1,362 from synthetic) identities. In the second stage, we replace the classification layer with a new classifier that will be trained on the real dataset (recognizing 333 classes). Table~\ref{table:two-stage} shows significant improvements with this method.

\begin{table}[t]
\centering
\caption{Re-ID test accuracy (mAP) on VehicleID test set (large) using various training datasets with~\cite{luo2019bag}. The first four training sets are generated by random attributes, random search, LTS and attribute descent, respectively. The last two training sets are real-world ones. FID measures domain gap between the training sets and VehicleID.} 
\resizebox{1\textwidth}{!}{
\begin{tabular}{l|cccccc} 
\Xhline{1.2pt}
& Ran. Attr. & Ran. Sear. & LTS  & Attr. Desc. & VeRi & Cityflow   \\ 
\hline
FID           &    134.75               &        109.94       &          95.27            &        77.96            &   -    &  -     \\ 
\hline
mAP           &    22.00         &       26.35         &          32.21             &    35.33               &    38.59 &   45.57      \\
\Xhline{1.2pt}
\end{tabular}}

\label{table:method_comp}
\end{table}

 \begin{table}[t]
\small
\caption{Method comparison when testing on VeRi-776. Both VehicleX training and joint training results are included. ``R'' means training with real data only, ``S'' represents training use synthetic data only and ``R+S'' denotes the joint training. VID$\rightarrow$VeRi shows the result trained on VehicleID, test on VeRi and Cityflow$\rightarrow$VeRi means the result trained on Cityflow, test on VeRi. In addition to some state-of-the-art methods, we summarize the results on top of two baselines, \emph{i.e.,} IDE~\cite{zheng2016mars} and PCB~\cite{sun2018beyond}.}
\resizebox{1\textwidth}{!}{
\begin{tabular}{l|r|c|ccc} 
\Xhline{1.2pt}
Experiment                   & Method                       & Data & Rank-1          & Rank-5          & mAP              \\ 
\hline
\multirow{5}{*}{VehicleX Training} & ImageNet                     & R    & 30.57           & 47.85           & 8.19             \\
                             & VID $\rightarrow$VeRi        & R    & 59.24           & 71.16           & 20.32            \\
                             & Cityflow $\rightarrow$ VeRi  & R    & 69.96           & 81.35           & 26.71            \\ 
\cline{2-6}
                             & Ran. Attr.                 & S    & 43.56           & 61.98           & 18.36            \\
                             & Attr. Desc.                & S    & 51.25           & 67.70           & 21.29            \\ 
\Xhline{1.2pt}
\multirow{7}{*}{Joint Training}   & VANet    ~\cite{chu2019vehicle}                   & R    & 89.78           & 95.99           & 66.34            \\
                             & AAVER   ~\cite{khorramshahi2019dual}                    & R    & 90.17           & 94.34           & 66.35            \\
                             & PAMTRI   ~\cite{tang2019pamtri}                   & R+S  & 92.86           & 96.97           & 71.88            \\ 
\cline{2-6}
                             & IDE                          & R    & 92.73           & 96.78           & 66.54            \\
                             & Ran. Attr.                 & R+S  & 93.21           & 96.20           & 69.28            \\
                             & Attr. Desc.                & R+S  & 93.44           & 97.26           & 70.62            \\ 
\cline{2-6}
                             & Attr. Desc. (PCB)           & R+S  & \textbf{94.34}  & \textbf{97.91}  & \textbf{74.51}   \\
\Xhline{1.2pt}
\end{tabular}}
\label{table:comparison-VeRi}
\end{table}

\subsection{Evaluation}\label{sec:single_cam}


\textbf{Analysis of attribute descent process.} Figure~\ref{fig:training} shows how the re-ID accuracy and FID change during along the training iterations. We observe that attributes are successively optimized when FID decreases and mAP increases. 
Furthermore, from the slope of the FID curve we can see that 
orientation has the largest impact on the distribution difference and mAP, with a huge FID drop from 147.85 to 91.14 and large mAP increase from 12.1\% to 21.94\%. Lighting is the second most impactful (-7.2 FID, +10.7\% mAP), and camera attributes are the third (-4.11 FID, +2.4\% mAP). 

\begin{figure}[t]
\centering
\includegraphics[width=0.72\textwidth]{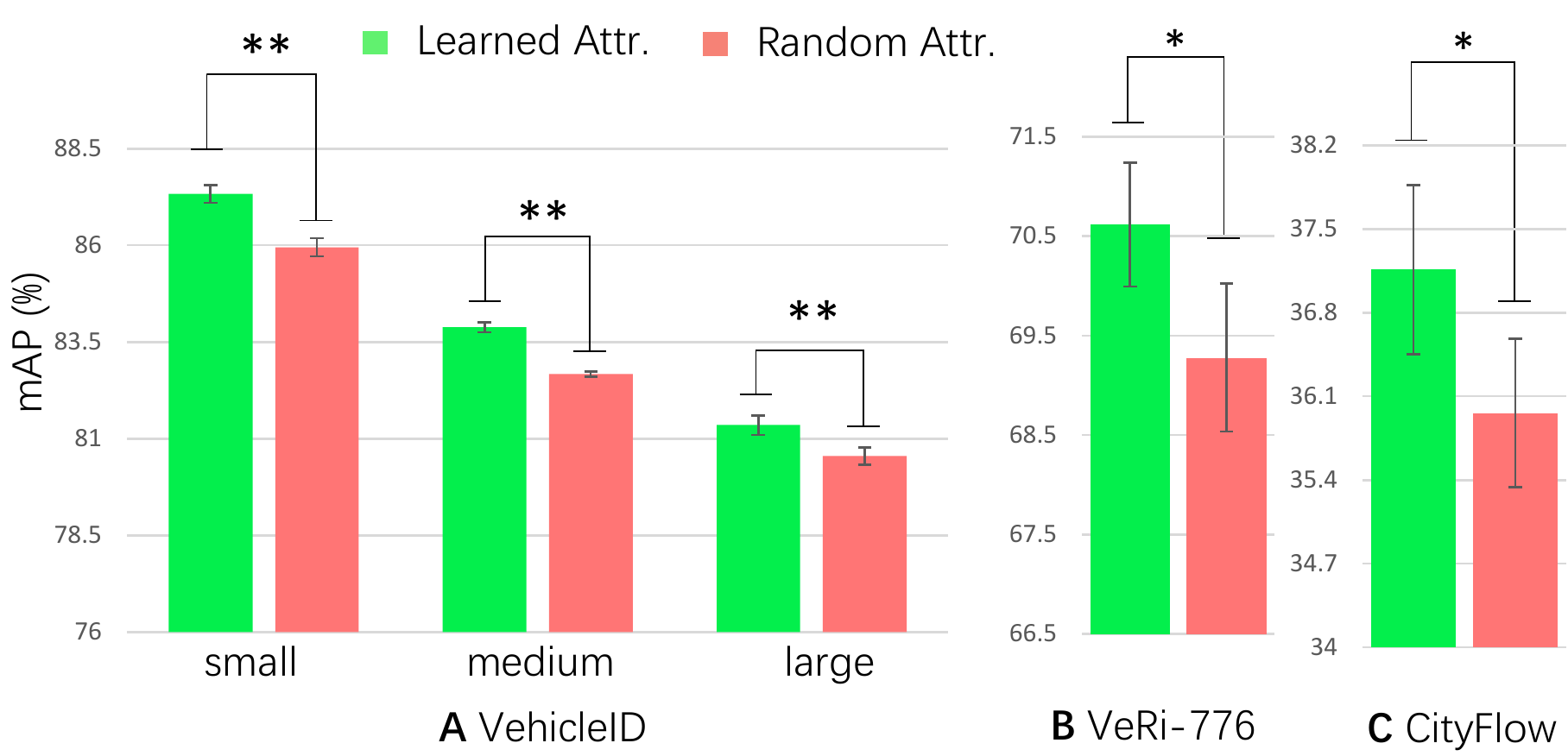}
\caption{Performance comparison between learned attributes and random attributes in joint training. We present mAP on three datasets and use statistical significance analysis to show the training stability.
$*$ means {statistically significant} ($i.e., 0.01 < p$-value $< 0.05$) and $**$ denotes {statistically very significant} ($i.e., 0.001 < p$-value $< 0.01$).} 
\label{figure: stas_analy}
\end{figure}

\textbf{Effectiveness of learned synthetic data.} Learned synthetic data can be used as a training set alone, or in conjunction with real training data for data augmentation. We show the results of both cases on the three datasets in Table~\ref{table:comparison-VID} (VehicleID), Table~\ref{table:comparison-VeRi} (VeRi) and Table~\ref{table:comparison-AIC} (CityFlow). 
From these results we observe that when used as training data alone, learned attributes achieve much higher re-ID accuracy than random attributes. For example, on the VeRi-776 dataset, attribute descent has a +7.69\% improvement in Rank-1 accuracy over random attributes. 
Moreover, attribute learning also benefits the data augmentation setting. For example, on CityFlow and VeRi, the improvement of learned attributes over random attributes is +1.49\% and +3.87\% in Rank-1 accuracy, respectively. Although this improvement looks small in number, we show that the improvement is statistical significant (Figure~\ref{figure: stas_analy}). 
We note that the improvement of using synthetic data as a training set is more significant than for data augmentation. When the training set consists of only the synthetic data, a higher quality of attributes will have a more direct impact on the re-ID results. 


\textbf{Few dependencies between attributes.} We proceed study on dependency by testing whether the order of attributes matters.
From Table~\ref{table:attr_dep} it is clear that attribute orders do not affect the downward trend, the only clear dependency is the relationship between orientation and camera. If we learn camera attributes before orientation, the accuracy will be influenced. But such influence will be eliminated by performing the attribute descent twice. Based on the few dependencies between attributes, we make it possible to use attribute descent rather than grid search, saving computation time.

\textbf{Attribute descent performs better than multiple methods: 1) random attribute 2) random search 3) LTS~\cite{ruiz2019learning}.} For LTS, we follow their ideas but we replace the task loss with FID score,
\begin{SCtable}
\resizebox{7cm}{1.4cm}{
\begin{tabular}{p{2.2cm}<{\centering}|p{0.6cm}<{\centering}|p{0.6cm}<{\centering}p{0.8cm}<{\centering}p{0.6cm}<{\centering}}
\Xhline{1.2pt}
Method  & Data  & R-1    & R-20    & mAP  \\
                             \hline
BA~\cite{kumar2019vehicle} &    R                               & 49.62    & 80.04  & 25.61    \\
BS~\cite{kumar2019vehicle}  &    R                              & 49.05    & 78.80  & 25.57    \\
PAMTRI~\cite{tang2019pamtri} &    R+S                                & 59.7&	80.13  & 33.81 \\\hline
IDE(CE+Tri.)&  R         & 56.75&	72.24   &  30.21 \\
Ran. Attr.  &  R+S    & 63.59    &  82.60    & 35.96    \\
Attr. Desc.  & R+S    &  \textbf{64.07}    & \textbf{83.27}   & \textbf{37.16}  \\
\Xhline{1.2pt}
\end{tabular}}
\caption{Method comparison on CityFlow with joint training. Our baseline is built with a combination of the CE loss and the triplet loss~\cite{luo2019bag}. Rank-1, Rank-20 and the mAP are calculated by the online server.}\label{table:comparison-AIC}
\end{SCtable}
since task loss is not generalised to a re-ID task. Our reproduced LTS uses the same distribution definition and initialization as attribute descent. In order to make a fair comparison, we report values from 200 iterations of training (\emph{i.e.,} compute FID score 200 times). Random search is a strong baseline in hyper-parameter optimization~\cite{bergstra2012random}.
In practice, we randomly sample attribute values 200 times and choose an attribute list with the best FID score. The result comparison is shown in Table~\ref{table:method_comp}. First, under the same task network, all learned attributes (\emph{i.e., random search, LTS and attribute descend}) perform better than random attributes, in both FID and mAP, showing that learned attributes significantly improves the result, and that content differences matter. Second, random search does not perform well in a limited search time. It has been shown that random search performs well when there exists many less important parameters~\cite{bergstra2012random}. But in our search space, all attributes contribute to the distribution differences as shown in Figure~\ref{fig:training}, thus random search has no advantage in helping find important attributes. Third, LTS works but it does not find a better FID score than attribute descent. LTS seems to fall into a local optimum and does not reach a global one. 
A example of local optima is LTS outputs are either outputs of car front or rear, whereas the VehicleID contains both car front and rear. With a more hand-crafted design, we will definitely reach a better performing LTS framework. But at this stage, attribute descent is a simple realized method that finds a better solution with few iterations. It deserves to be a strong baseline in this field.  

\textbf{Comparison with the state-of-the-art.} When the synthetic data is used in conjunction with real-world training set, we achieve very competitive accuracy compared with the state-of-the-art (Table~\ref{table:comparison-VID}, Table~\ref{table:comparison-VeRi} and Table~\ref{table:comparison-AIC}). For example on VehicleID (Small), our method is +5.6\% higher than~\cite{bai2018group} in 
Rank-1 accuracy. On CityFlow, our method is higher than~\cite{tang2019pamtri} by +7.32\% in Rank-1 accuracy. 

\section{Conclusion}
This paper study the domain gap problem
between synthetic data and real data
from the content level. That is, we automatically edit the source domain image content in a graphic engine so as to reduce the content gap between the synthetic images and the real images. We use this idea to study the vehicle re-ID task, where the usage of vehicle bounding boxes decreases the set of attributes to be optimized. Fewer attributes-of-interest and low dependencies between them allow us to optimize them one by one using our proposed attribute descent approach. We show that the learned attributes bring about improvement in re-ID accuracy with statistical significance. Moreover, our experiment reveals some important insights regarding the usage of synthetic data, \emph{e.g.,} style DA brings significant improvement and two stage training is beneficial for joint training. 

\section*{Acknowledgement}
Dr. Liang Zheng is the recipient of Australian Research Council Discovery Early Career Award (DE200101283) funded by the Australian Government.

\clearpage
%
%
\bibliographystyle{splncs04}
\bibliography{egbib}
\end{document}